\documentclass[sigconf, nonacm]{aamas} 
\usepackage{balance} 
\usepackage{graphicx}  
\usepackage{epstopdf}  

\usepackage{multirow}  
\usepackage[table]{xcolor}  
\usepackage{graphicx}                
\usepackage{booktabs}                
\usepackage[most]{tcolorbox}
\definecolor{t2e_red}{RGB}{239,99,75}
\definecolor{t2e_blue}{RGB}{99,113,250}
\definecolor{t2e_green}{RGB}{0,180,139}
\definecolor{t2e_gray}{RGB}{165,165,165}
\definecolor{redlink}{RGB}{239,99,75}
\usepackage{dblfloatfix}   
\usepackage{booktabs}
\usepackage[table]{xcolor}
\usepackage{multirow}
\usepackage{rotating} 
\usepackage{tabularx,array,booktabs}
\usepackage{colortbl}

\newcolumntype{L}{>{\raggedright\arraybackslash}X}
\newcolumntype{R}[1]{>{\raggedleft\arraybackslash}p{#1}}

\usepackage[table]{xcolor} 
\newcommand{\ulshade}[2][gray!20]{\cellcolor{#1}{#2}} 

\usepackage{siunitx}
\usepackage{pifont}
\newcommand{\cmark}{\ding{51}}
\newcommand{\xmark}{\ding{55}}

\title[AAMAS-2026 Formatting Instructions]{RoboGPT-R1: Enhancing Robot Task Planning with Reinforcement Learning}

\author{Jinrui Liu}\authornote{Equal contribution.}
\affiliation{
  \institution{Institute of Automation, CASIA}
  \institution{School of Artificial Intelligence, UCAS}
  \city{Beijing}
  \country{China}}
\email{liujinrui2024@ia.ac.cn}

\author{Bingyan Nie}\authornotemark[1]
\affiliation{
  \institution{Institute of Automation, CASIA}
  \institution{School of Artificial Intelligence, UCAS}
  \city{Beijing}
  \country{China}}
\email{niebingyan2025@ia.ac.cn}

\author{Boyu Li}
\affiliation{
  \institution{Institute of Automation, CASIA}
  \institution{School of Artificial Intelligence, UCAS}
  \city{Beijing}
  \country{China}}
\email{liboyu2021@ia.ac.cn}

\author{Yaran Chen}
\affiliation{
  \institution{Institute of Automation, CASIA}
  \institution{School of Artificial Intelligence, UCAS}
  \city{Beijing}
  \country{China}}
\email{chenyaran2013@ia.ac.cn}

\author{Yuze Wang}
\affiliation{
  \institution{Huawei Cloud Technology Co., Ltd}   
  \city{Shenzhen}
  \country{China}}
\email{wangyuze1@hisilicon.com}            

\author{Shunsen He}
\affiliation{
  \institution{Huawei Cloud Technology Co., Ltd}  
  \city{Shenzhen}
  \country{China}}                      
\email{heshunsen@huawei.com}            

\author{Haoran Li}\authornote{Corresponding author.}
\affiliation{
  \institution{Institute of Automation, CASIA}
  \institution{School of Artificial Intelligence, UCAS}
  \city{Beijing}
  \country{China}}
\email{lihaoran2015@ia.ac.cn}

\begin{abstract}
 Improving the reasoning capabilities of embodied agents is crucial for robots to complete complex human instructions in long-view manipulation tasks successfully. Despite the success of large language models and vision language models based on Supervised Fine-Tuning (SFT) in planning tasks, they continue facing challenges in performing long-horizon manipulation tasks in complex real-world environments, owing to their restricted common sense and reasoning capabilities. 
 Considering that aligning general-purpose vision language models to robotic planning tasks via supervised fine-tuning suffers from poor generalization and insufficient physical understanding, we propose RoboGPT-R1, a two-stage fine-tuning framework for embodied planning. In this framework, supervised training acquires foundational knowledge through expert sequences, followed by RL to address the model's shortcomings in visual-spatial understanding and reasoning. To achieve physical understanding and action sequence consistency in multi-step reasoning tasks, we design a rule-based reward function that simultaneously considers long-horizon performance and action constraint in the environment. The reasoning model, trained on Qwen2.5-VL-3B, significantly outperforms the larger-scale model, GPT-4o-mini, by 21.33\% and surpasses other work trained on Qwen2.5-VL-7B by 20.33\% on the EmbodiedBench benchmark. All code and details will be released at \url{https://github.com/Jinrui-Liu137/RoboGPT-R1}.
\end{abstract}


\newcommand{\BibTeX}{\rm B\kern-.05em{\sc i\kern-.025em b}\kern-.08em\TeX}

\begin{document}


\pagestyle{fancy}     
\fancyhead{}          


\maketitle 


\section{Introduction}
Recently, vision language models (VLMs) have been increasingly employed as high-level planners for embodied tasks~\cite{RewardMap,AlphaDrive,real-rl}, given their emerging capability to ground natural language instructions into long-horizon robotic action sequences. Nevertheless, in real-world environments, VLMs still fail to meet the demands of robustness and generalization~\cite{VADv2,RLVMR,real-rl}. Two major challenges remain. First, the prevailing supervised fine-tuning (SFT)-only paradigm primarily imitates expert demonstrations, yet lacks mechanisms for adaptation or self-correction in dynamic environments~\cite{sft-memory, BCVLR}. Second, the design of long-horizon reward functions remains inadequate—existing rewards are often sparse or poorly aligned with the execution of grounded action, ultimately hindering planning performance~\cite{REBP}.

In real-world long-horizon embodied tasks, VLMs still exhibit limited planning capability~\cite{REBP}, as they are not well aligned with the physical realities of robotic embodiments or with accurate state transition dynamics~\cite{vlm-robot,real-rl}. Although existing approaches based on the SFT-only paradigm can enhance the performance of VLMs~\cite{Embodied-Reasoner,RoboGPT}, they remain ineffective when confronted with scenarios or instructions that fall outside the distribution of the SFT dataset~\cite{EmbodiedBench}. The lack of physical common sense and feasibility constraints often results in ambiguous object recognition and biased state estimation~\cite{survey:embodiedai,Large-Model-survey}. Moreover, the absence of feedback and error-correction signals~\cite{rise:embodied-model} in the SFT paradigm encourages models to memorize answers rather than learn generalizable reasoning strategies~\cite{sft-memory}, thereby failing to mitigate the accumulation of local errors over extended task horizons.

Reinforcement learning (RL) has proven effective for VLMs in domains such as video reasoning~\cite{VideoChat-R1}, object detection~\cite{VLM-R1,Visual-RFT}, and mathematical reasoning~\cite{math,math2,GRPO-MA}, where tasks provide clear and verifiable answers. However, when transferred to open-ended embodied planning tasks, RL-based methods face challenges in designing dense and interpretable reward functions~\cite{AlphaDrive,RLVMR,vlm-survey}, as the outcomes are often ambiguous and context-dependent. For instance, in embodied planning tasks, when the reference plan is "pick up an apple and put it on the table", using a straightforward RL reward such as string matching or accuracy calculation allows the model to gain higher rewards simply by generating more actions, as some subsequences are likely to overlap with the reference plan. This mechanism misleads the model to produce overly long yet logically incorrect reasoning chains, masking its true deficiencies in action ordering and planning coherence. Therefore, a dense and sequence-aware reward is required to directly capture whether a multi-step plan is executed fully or partially correctly~\cite{vlm-robot,real-rl,vlm-survey}, particularly in long-horizon and complex action sequences, rather than rewarding superficial token overlap.

To address the above problems, we propose RoboGPT-R1, a two-stage training framework designed to enhance robotic planning with small-scale models. In contrast to the SFT stage, which learns predefined answers, the Group Relative Policy Optimization (GRPO) algorithm explores optimal solutions, addressing the shortcomings of SFT in generalization, task understanding, spatial perception, and planning consistency~\cite{sft-memory}. Moreover, in second-stage RL training, unlike conventional RL approaches in reasoning tasks that typically rely on sparse or single-point accuracy rewards, our method introduces a rule-based variable reward function specifically designed for long-horizon embodied reasoning and planning. This reward function consists of two complementary components: a format reward and an accuracy reward. As illustrated in Fig.~\ref{fig:framework}, the format reward integrates multiple dimensions, including structural completeness of reasoning, action type correctness, and action validity. The accuracy reward is based on the longest common subsequence (LCS) between predicted and reference action sequences, effectively preserving action order and enhancing long-horizon performance.

The results on EmbodiedBench~\cite{EmbodiedBench} show that RoboGPT-R1 significantly outperforms GPT-4o-mini and is competitive with closed-source models such as GPT-4o and Gemini-2.0-flash. Furthermore, its performance surpasses open-source models like Llama-3.2-90B, achieving a 23.33\% higher overall score. Compared to the previous state-of-the-art~\cite{REBP}, it yields a 20.33\% improvement. On long-horizon tasks, it leads with an accuracy of 50\%, demonstrating the superior reasoning capabilities of small models.

In summary, our contributions are as follows:
\begin{itemize}
\item We propose RoboGPT-R1, a two-stage training paradigm for embodied multi-step reasoning tasks. With RL training, RoboGPT-R1 develops the reasoning capability in complex tasks and environments, thereby enhancing its physical commonsense and error correction abilities. 
\item We design a reward function based on perception-reasoning-planning-action loop, with LCS reward effectively enhancing the model's understanding and self-correction. This enables efficient and high-quality reward computation at a very low cost and demonstrates good reasoning capabilities on long-horizon tasks.
\item We conduct extensive experiments on 6 tasks in 2 scenarios, including spatial perception, long-horizon reasoning, common-sense questions, and visual understanding. In seen scenarios, our method outperforms open-source general models and existing embodied planning models, achieving competitive performance compared to closed-source general models. Moreover, in unseen scenarios, it demonstrates superior reasoning ability and surpasses the state-of-the-art embodied planning model.
\end{itemize}


\section{Related Work}
\subsection{Embodied Planning}
Embodied agents require not only active exploration, manipulation, and scene perception, but also embodied task planning capabilities~\cite{TAPA,survey:agentai,survey:embodiedai,SeqWM}. Embodied planning aims to decompose high-level natural language instructions into executable subtask sequences~\cite{Socratic-Planner,Unleashing}, enabling the embodied agent to generate actionable steps within an interactive environment to complete complex behaviors. With the advent of large language models~\cite{rise:embodied-model,rise:llm,survey:embodiedai2}, natural language offers greater expressive flexibility than structured languages, making it possible to utilize LLMs to decompose complex plans into sub-plans in a fully automated manner~\cite{RLVMR, llm-limit, DigiRL:llm, ScienceWorld:llm, DipLLM}. For example, TaPA introduces an embodied task planner that grounds free-form instructions into executable plans, trained on a new multimodal benchmark (80 scenes, 15K instruction–plan pairs)~\cite{TAPA}. It fine-tunes LLaMA with object lists from multi-view open-vocabulary detection. Additionally, SayCan~\cite{saycan} combines an LLM with reinforcement learning, leveraging the high-level reasoning capabilities of LLM to complement the value assessment of pre-trained skills, thereby laying a foundation for language in robotics and enabling the feasible scoring of actions. This can generate executable long-term plans suitable for real robots. While LLMs can generate preliminary plans based on commonsense reasoning, they lack constraints on the physical environment and the feasibility of actions~\cite{llm-limit,llmplanner,voyager,Grounded-decoding}. The emergence of VLMs~\cite{Hi-Robo,TAPA,vlm-robot} has led to their use as high-level planners, with the current mainstream approach being to fine-tune VLMs based on demonstration data. Zhang et al.~\cite{RLVMR} extend the O1-style deep reasoning to embodied interactive tasks by coupling visual search with step-by-step planning, reflection, and verification, trained on synthesized Observation–Thought–Action trajectories to improve performance on AI2–THOR–style~\cite{AI2-THOR} tasks. Moreover, Reflective Planning~\cite{Reflective-Planning} proposes a test-time computation framework that augments a pre-trained VLM with a reflection loop. It imagines future world states via a diffusion-based dynamics model, critiques potential suboptimalities, and revises plans to improve multi-stage, long-horizon manipulation. 
\subsection{Reinforcement Learning for LLMs and VLMs}
In recent years, with the emergence of reasoning models like OpenAI's o1~\cite{openaio1}, research on large language models (LLMs) has gradually shifted towards enhancing their reasoning capabilities through reinforcement learning (RL)~\cite{SRFT,llm-reason,SWE-RL,tsj-llmdpo,tsj-onlinerl,CriticSearch}. Numerous studies have explored ways to enhance the performance of LLMs in reasoning tasks, including solving mathematical problems~\cite{math2,deepseekmath,tsj-r1} and coding~\cite{acecoder}. A notable breakthrough in this field is DeepSeek-R1~\cite{deepseek-r1}, which experienced an "aha moment" during GRPO-based training, enabling the model to independently reflect and reevaluate its initial policy without any explicit guidance. Subsequently, several works~\cite{VLM-R1,Visual-RFT,kimi,VideoChat-R1,R1-VL,MM-Eureka} have used reinforcement learning to enhance the reasoning capabilities of models in multimodal settings. For example, VLM-R1~\cite{VLM-R1} and Visual-RFT~\cite{Visual-RFT} extend R1-style reinforcement learning to vision-language models, sampling multiple answer outputs for each input and optimizing with GRPO using verifiable, task-specific rewards, resulting in stronger visual reasoning and perception enhancements over the SFT baseline. These advances demonstrate the potential of RL to propel large models from "imitation learning" to "emergent intelligence"~\cite{sft-memory}. Inspired by the R1 paradigm, this paper employs GRPO-based reinforcement learning to perform two-stage training on the model, systematically improving the planning ability and long-term consistency of embodied agents in multimodal task planning.

\section{Methodology}

\subsection{Overview}
\begin{figure*}[ht] %
  \centering
  \includegraphics[width=0.9\linewidth, height=9cm]{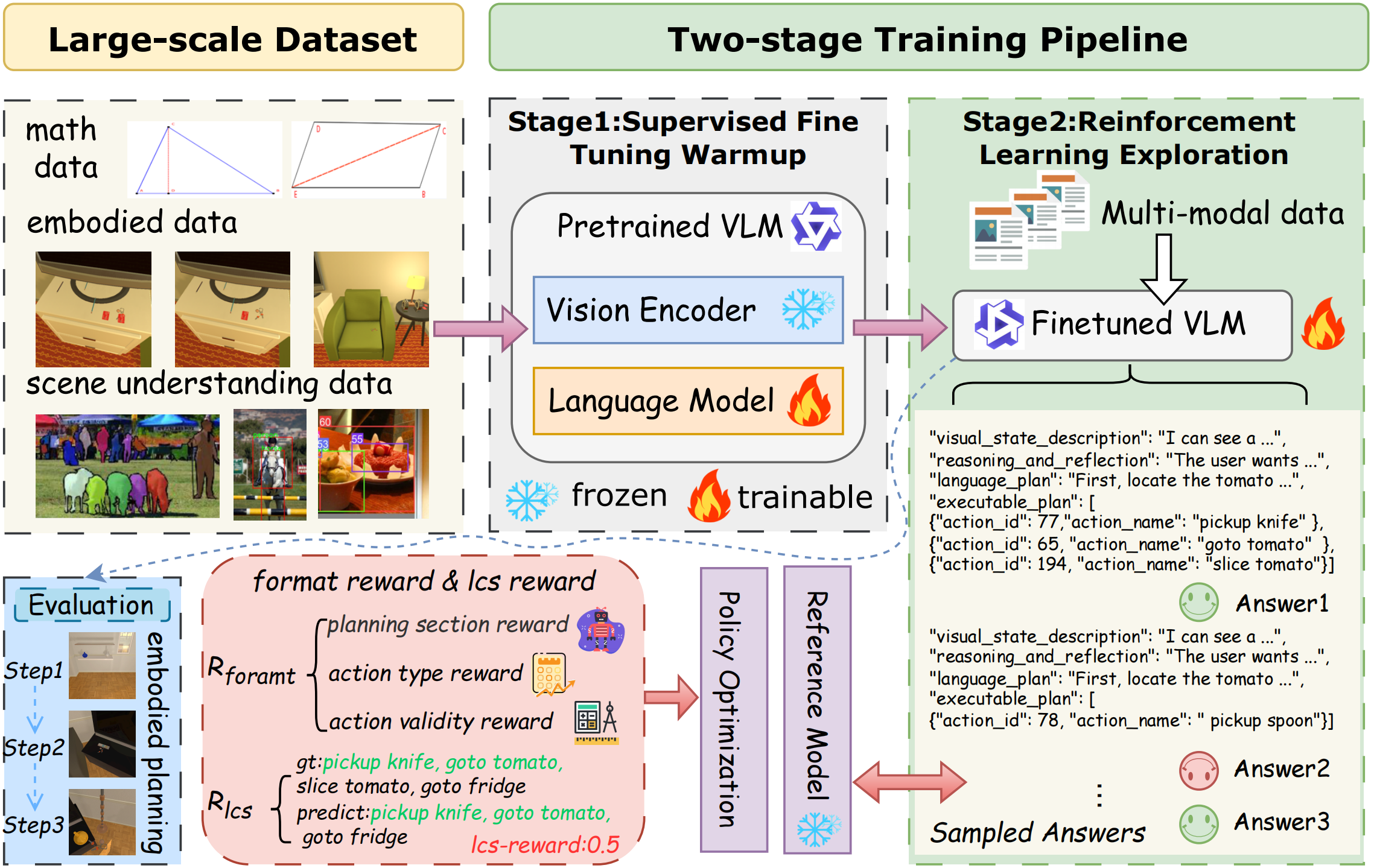}
  \Description{Diagram of the RoboGPT-R1 framework, showing a large-scale multimodal dataset and a two-stage training pipeline (supervised fine-tuning warmup followed by reinforcement learning) with reward-guided optimization and evaluation on embodied planning tasks.}
  \caption{An overview of RoboGPT-R1. RoboGPT-R1 adopts a two-stage learning paradigm. In the initial phase, supervised fine-tuning introduces the model to data from mathematics, embodied tasks, and visual understanding, establishing a foundation for embodied reasoning. In the second phase, we apply GRPO-based reinforcement fine-tuning guided by a tailored reward function. The model is subsequently evaluated across six categories of tasks, including long-horizon planning and spatial reasoning. }
  \label{fig:framework}
\end{figure*}

In this section, we provide a brief introduction to the proposed RoboGPT-R1 framework. In contrast to previous approaches that solely rely on SFT, this study explores the incorporation of RL and reasoning techniques to better align the model with embodied planning tasks. Section~\ref{sec:training_scheme} will introduce the two-stage learning scheme. As demonstrated in Fig.~\ref{fig:framework}, the training of agents is comprised of two phases: an initial SFT phase, the purpose of which is to instill fundamental knowledge and elementary reasoning capabilities into the agent; and a subsequent reinforcement learning phase, the function of which is to utilize the GRPO policy optimization algorithm to enable the agent to continually explore, think, and learn independently. Next, to address issues such as unstable step-to-step coherence in multi-step reasoning, limited error recovery, and poor performance on long-term tasks, we designed a rule-based, verifiable reward to incentivize the VLM's planning capabilities in Section \ref{sec:reward_design}.

\subsection{Data Preparation}
Embodied task planning in real-world indoor scenarios requires a substantial amount of multimodal data, encompassing both perceptual and physical knowledge. Given the impressive inference performance of large, closed-source models, data distillation can be employed to generate high-quality datasets. Following REBP~\cite{REBP}, we employ the SFT dataset distilled from Gemini-2.0-flash, in the SFT phase. In the RL phase, the RFT dataset from REBP is augmented with task-relevant examples and unsuccessful exploratory tasks. In our preliminary studies, we observed that the quantity of examples contained within a dataset can result in contradictory outcomes regarding the training and testing of models. For instance, the direct application of n-shot (n denotes the number of examples) to the SFT phase results in the model overfitting to the provided demonstrations. However, during testing, there is a significant drop in the n-shot score. In contrast, an untrained model significantly improves its score when tested using n-shot, demonstrating its firm reliance on examples during the testing process. Consequently, it can be concluded that the injection of knowledge into a trained model may result in the model's rigid adherence to predefined answer templates. This phenomenon impedes the model's capacity to adapt to the complexity of the problem and the variability of the environment. Experiments have demonstrated that zero-shot training and testing models exhibit superior performance, while concurrently reducing the number of input tokens by approximately one-third (from 9,000+ to less than 6,000), thereby enhancing the efficiency of the training process. Consequently, we employ zero-shot processing uniformly for both training and testing data.

\subsection{Two-stage Training Scheme}
\label{sec:training_scheme}
\textbf{Stage 1: Initial Planning via SFT.} To equip the base VLM with the fundamental capacity to generate multi-step reasoning, it is first necessary to undertake a supervised fine-tuning phase. This step is crucial because the reasoning patterns learned in subsequent reinforcement learning are significantly affected by the capabilities of the base model. Furthermore, the use of reinforcement learning as the sole training method is found to be significantly affected by data distribution, resulting in instability during the initial training stages. Therefore, we use a small amount of the data for an SFT-based warm-up stage. Following this, reinforcement learning training is conducted using the entire dataset. The findings of this study demonstrate that this approach enhances stability in the initial stages of training and rapidly integrates the base model with relevant knowledge, thereby enabling it to acquire a certain level of embodied reasoning knowledge and planning capabilities.

\noindent\textbf{Stage 2: Enhancing reasoning with GRPO.} The DeepSeek R1-zero algorithm employs the GRPO framework. Unlike reinforcement learning algorithms such as PPO, which require an additional critic model to estimate policy performance, GRPO directly compares groups of candidate responses, eliminating the need for a separate critic. Given a question $q$, GRPO samples $N$ candidate responses $\left \{ o_1, o_2,\dots , o_N \right \}$ from the policy $\pi _{\theta} $ and evaluates the quality of each response $o_i$ using a reward function $R\left( q, o_i \right) $. To determine the relative quality of these responses,  the algorithm normalizes the rewards by computing their mean and standard deviation and subsequently derives the advantage as:
\begin{equation}
A_i = \frac{r_i - \operatorname{mean}(\{r_1, r_2, \dots,r_N\})}{\operatorname{std}(\{r_1, r_2,\dots,r_N\})}
\end{equation}
where $A_i$ denotes the advantage of candidate $o_i$ measured against the other samples in the group. To encourage the model to generate responses with higher advantages within the group, the group penalizes large deviations from the reference policy $\pi _{ref } $. The objective is as follows:

\begin{equation}
\begin{aligned}
\mathcal{J}_{\text{grpo}}(\theta) =\; & \mathbb{E}_{q, \{o_i\}_{i=1}^N \sim \pi_{\theta_{\text{old}}}(q)} \\
& \Bigg[ \frac{1}{N} \sum_{i=1}^{N} 
\left\{ \min\left( \ell_1 \cdot A_i,\; \ell_2 \cdot A_i \right) 
- \beta D_{\text{KL}}\left( \pi_\theta \,\|\, \pi_{\text{ref}} \right) \right\} \Bigg]
\end{aligned}
\end{equation}

\begin{equation}
\ell_1=\frac{\pi_\theta(o_i \mid q)}{\pi_{\theta_{\text{old}}}(o_i \mid q)};
\ell_2= \operatorname{clip} \left( \frac{\pi_\theta(o_i \mid q)}{\pi_{\theta_{\text{old}}}(o_i \mid q)}, 1 - \epsilon, 1 + \epsilon \right) 
\end{equation}
where  $\epsilon$ is the clipping hyperparameter and $\beta$ controls the KL penalty against a reference policy $\pi_{\text{ref}}$.
Inspired by DeepSeek-R1, our approach incorporates two complementary rewards: the accuracy reward and the format reward. Next, we introduce the reward function $R$ designed for robotic planning tasks.

\subsection{Reward Design}
\label{sec:reward_design}
In Section~\ref{sec:training_scheme}, we have briefly introduced the general GRPO algorithm. Given that embodied planning requires an agent to complete complex tasks in a real or simulated physical environment based on natural language instructions, we design a set of reward functions specifically for this purpose. The following sections describe the format reward and the accuracy reward, respectively.

\subsubsection{Format Reward}
To ensure output regularity and facilitate the extraction of the reasoning process and the final result, most R1-style approaches enclose the reasoning within `\texttt{<think></think>}' tags and the final plan within `\texttt{<answer></answer>}' tags. If the generated output deviates from this structure, the format reward is assigned a value of zero. Inspired by REBP~\cite{REBP}, in embodied multi-step planning, the agent must generate responses that are not only semantically meaningful but also structurally executable. Unlike conventional text generation tasks, where free-form output may be acceptable, embodied planning requires a higher level of structural rigour to ensure that each response remains interpretable and executable by downstream systems. First, the model should follow a clear cognitive loop—just like how humans plan their actions. Before acting, we reflect on the task, observe the environment, make a plan, and then execute it. Second, the generation of invalid or fabricated actions should be penalized. In summary, we set the following reward format, which consists of three parts:
\begin{equation}
R_{\text{format}} = 0.3 \cdot R_{\text{section}} + 0.3 \cdot R_{\text{type}} + 0.4 \cdot R_{\text{validity}}
\end{equation}

The section reward $R_{\text{section}}$ evaluates whether all required fields   (\nolinkurl{visual_state_description},
\nolinkurl{reasoning_and_reflection},
\nolinkurl{language_plan},
\nolinkurl{executable_plan}) are present and correctly typed, and is computed as:
\begin{equation}
R_{\text{section}} = \frac{1}{|S|} \sum_{s \in S} \mathbf{1}\left[ \text{type}(o_s) = T_s \right]
\end{equation}
where $\mathbf{1}$ is an indicator function, $S= \left \{ s_1,s_2,s_3,s_4 \right \} $ includes the four fields,$T_s$ denotes the expected type of each field and $o_s$ is the value of field $s$ in the output object. The type reward  $R_{\text{type}}$ checks whether each action step is well-formed, defined as:
\begin{equation}
R_{\text{type}} = \frac{1}{m} \sum_{i=1}^{m} \mathbf{1}\left[ \hat{y}^{id}_i  \in \text{Int} \; \wedge\ \; \hat{y}^{name}_i \in \text{Str} \right]
\end{equation}
In this formulation, $m$ denotes the number of action steps, $\hat{y}^{id}_i$ and $\hat{y}^{name}_i$ represent the action id and name of step $i$ respectively. Here $\text{Int}$ denotes the set of integers, while $\text{Str}$ represents the set of non-empty strings. Finally, the validity reward $R_{\text{validity}}$ measures whether each action id–name pair matches the predefined action dictionary, given by:
\begin{equation}
R_{\text{validity}} = \frac{1}{|C|} \sum_{i \in C} \mathbf{1}\left[ \text{norm}(\hat{y}^{name}_i) = \text{norm}(\mathcal{D}_{\text{action}}[\hat{y}^{id}_i]) \right]
\end{equation}
where $C$ is the index set of steps whose action ids are defined in the action dictionary $\mathcal{D}_{\text{action}}$ and $\text{norm}(\cdot)$ is a normalization function that lowercases and trims whitespace. The action dictionary $\mathcal{D}_{\text{action}}$ defines a mapping from action ids to their corresponding names.

Unlike REBP~\cite{REBP}, our setup employs dynamic action IDs that vary across tasks and environments, preventing the model from relying on memorization of a fixed action set. Instead, we want it to learn the meaning of actions and use them correctly. In summary, such a reward design can guide the model to generate structured outputs that conform to the task execution closed loop, and avoid hallucinations or inconsistent outputs through self-understanding and thinking. 

\subsubsection{LCS Reward}
In embodied multi-step planning, the correctness of individual actions is not sufficient—the order in which actions are executed is often critical to task success. For example, placing an object before picking it up may involve the right actions but in a logically invalid sequence. Traditional token-level matching or step-wise accuracy metrics fail to penalize such disorder, treating unordered but correct actions as equally valid. Besides, in long-horizon tasks, plans can extend over dozens of steps, where strict reward strategies like exact matching become too rigid to reflect realistic performance. A model might make early mistakes yet recover in later steps to complete the task successfully. However, prefix-based accuracy rewards, such as those used in REBP~\cite{REBP}, overlook this “error recovery” behavior, leading to sparse and less informative reward signals.
To address these problems, we design the accuracy reward based on the Longest Common Subsequence (LCS) between the predicted and reference action sequences. By computing the LCS over action names, we enforce both content accuracy and sequence coherence. This approach is robust to local deviations while maintaining global alignment and remains effective even as task length increases. We define the model generation sequence as 
$\hat{Y} = (\hat{y}_1, \hat{y}_2, \dots, \hat{y}_m)$ and the reference sequence as $\quad
Y = (y_1, y_2, \dots, y_n)$ . The detailed accuracy reward is as follows:
\begin{equation}
\text{LCS}(i, j) = 
\begin{cases}
0, & \text{if } i = 0 \text{ or } j = 0 \\
\text{LCS}(i-1, j-1) + 1, & \text{if } \hat{y}_i = y_j \\
\max\left( \text{LCS}(i-1, j),\; \text{LCS}(i, j-1) \right), & \text{otherwise}
\end{cases}
\end{equation}

\begin{equation}
R_{\text{lcs}} = \frac{k}{n},
\end{equation}
where $n$ is the length of the reference sequence $|Y|$, $k$ is the length of the longest common subsequence $\text{LCS}(m, n)$. 
To evaluate the effectiveness of our proposed accuracy reward, we conduct an ablation study. Compared to prefix accuracy (as used in REBP) and standard step-wise matching, the LCS-based accuracy reward shows stronger performance in multi-step  planning tasks.

\subsubsection{Overall Reward}
To jointly encourage structural correctness and sequential accuracy, we define the overall reward as a weighted combination of the format reward and the LCS-based accuracy reward:
\begin{equation}
 R = 0.2 \cdot R_{\text{format}}+0.8 \cdot R_{\text{lcs}}
\end{equation}
This overall reward formulation has two key purposes in embodied planning: enforcing structural correctness and promoting action-level accuracy. The format reward $R_{\text{format}}$ enforces a fixed output structure and ensures each action step is well-formed and consistent with the current scene and instructions. Meanwhile, the accuracy reward $R_{\text{lcs}}$ evaluates whether the predicted action sequence aligns with the reference plan, not only in content but also in order. By combining these two aspects, the overall reward encourages the model to reason more effectively and generate executable plans with reasonable length and structure.

\section{Experiments}
\label{sec:experiments}

\subsection{Experimental Settings}
\noindent\textbf{Evaluation.}
We evaluate RoboGPT-R1 in EmbodiedBench~\cite{EmbodiedBench}, a unified benchmarking suite for multimodal embodied planning agents. EmbodiedBench offers standardized protocols and interfaces that cover multi-sensory input, language instructions, and long-horizon decision making, enabling consistent and reproducible comparisons across task suites. Specifically, we focus on its two constituent suites: EB-ALFRED and EB-Habitat. The former is rooted in the ALFRED ecosystem and targets instruction-following household tasks (e.g., pick-and-place, cleaning, and mobile manipulation) that emphasize object state tracking and stepwise dependencies; the latter builds on the Habitat ecosystem and emphasizes navigation and interaction in 3D environments, with observation distributions, scene layouts, and action semantics that differ markedly from EB-ALFRED~\cite{ALFRED,Habitat}. Because our training data are primarily associated with EB-ALFRED/ALFRED, we treat performance on EB-ALFRED as in-domain and use it to gauge method effectiveness. Results on EB-Habitat are regarded as out-of-domain and used to assess generalization.


Unless otherwise noted, all evaluation settings follow the benchmark defaults; the only test variable we modify is the number of examples in context (\texttt{n\_shots}) in the control input. The evaluation setting details are provided in the appendix~\ref{app:eval}     


\noindent\textbf{Baselines.}
We evaluate both the Qwen and GPT model series, including Qwen2.5-VL-72B-Ins, Qwen2.5-VL-7B-Ins, Qwen2.5-VL-3B-Ins, as well as GPT-4.1, GPT-4o and GPT-4o-mini. Closed-source models are assessed via their official APIs, while open-source models are tested through local deployment. Performance results for additional baselines are obtained from the REBP and EmbodiedBench leaderboards. We group baselines into three categories, and present the corresponding results below:
\begin{enumerate}
\item \textbf{General closed-source models:} including five representative proprietary multimodal models from the GPT, Gemini, and Qwen families: Gemini-2.0-Flash~\cite{gemini2}, Qwen-VL-Max~\cite{qwenvlmax}, GPT-4.1~\cite{gpt4.1}, GPT-4o~\cite{gpt4o} and GPT-4o-mini~\cite{gpt4o-mini}.
\item \textbf{General open-source models:} Comprising open-source models from multiple series—including Qwen2.5‑VL‑3B/7B/\ 72B-Ins~\cite{qwen2.5vl}, LLaMA‑3.2‑90B‑Vision‑Ins~\cite{llama3.2}, InternVL2.5‑8B\ \ ~\cite{internvl2.5}, and Gemma‑3‑12B‑it~\cite{gemma3technicalreport}.
\item \textbf{Embodied domain-specific models:} focusing on models tailored for embodied reasoning and planning, such as REBP~\cite{REBP}, RoBoBrain~\cite{robobrain}, TaPa~\cite{TAPA} and ours.
\end{enumerate}


\begin{table*}[!t]
\centering
\caption{Success rates of diverse models on EB-ALFRED and EB-Habitat. Entries without any symbol are sourced from the EMBench leaderboard. Symbol $\dagger$ indicates results directly cited from the REBP paper, based on their evaluations. Symbol $\ddagger$ marks results obtained through our own reproduction by querying the official API. All scores are reported as percentages (\%). For each metric, the best-performing result is highlighted with a gray background.}
\label{tab:main_result}
\setlength{\tabcolsep}{3pt}
\resizebox{0.95\linewidth}{!}{%
\begin{tabular}{l|c|c|cccccc|c|cccccc}
\toprule
\multirow{2}{*}{\textbf{Method}}
& \multirow{2}{*}{\textbf{Params}}
& \multicolumn{7}{c|}{\textbf{EB-ALFRED (seen)}} 
& \multicolumn{7}{c}{\textbf{EB-Habitat (unseen)}} \\
\cmidrule(lr){3-9}\cmidrule(lr){10-16}
  & &\textbf{Avg.} & \textbf{Base} & \textbf{Common} & \textbf{Complex} & \textbf{Visual} & \textbf{Spatial} & \textbf{Long} &
\textbf{Avg.} & \textbf{Base} & \textbf{Common} & \textbf{Complex} & \textbf{Visual} & \textbf{Spatial} & \textbf{Long} \\
\midrule\midrule

\rowcolor{t2e_blue!15}\multicolumn{16}{l}{\textcolor{t2e_blue}{$\bullet$~\textbf{Type: Closed-Source General Model}}} \\

Gemini-2.0-flash~\cite{gemini2} 
& - & 52.30 & 62 & 48 & 54 & 46 & 46 & 58  
& 42.30 & 82 & 38 & 38 & 36 & 34 & 26 \\

Qwen-VL-Max~\cite{qwenvlmax} 
& - & 41.30 & 44 & 48 & 44 & 42 & 38 & 32
& 45.30 & 74 & 40 & 50 & \ulshade{42} & 30 & 36 \\

GPT-4.1$^{\ddagger}$~\cite{gpt4.1} 
& - & \ulshade{64.67} 
& \ulshade{70} & \ulshade{64} & \ulshade{70} 
& \ulshade{62} & \ulshade{62} & \ulshade{60}
& 50.67 & 90 & 38 & 50 & 36 & 46 & 44 \\

GPT-4o$^{\ddagger}$~\cite{gpt4o} 
& - & 51.67 & 54 & 46 & 58 & 52 & 52 & 48 
& \ulshade{57.00} & 84 & \ulshade{42} & \ulshade{62} & 38 & \ulshade{62} & \ulshade{54} \\

GPT-4o-mini$^{\ddagger}$~\cite{gpt4o-mini} 
& - & 34.00 & 66 & 56 & 68 & 14 & 0 & 0   
& 35.00 & 70 & 24 & 36 & 30 & 30 & 20 \\

\rowcolor{t2e_green!15}\multicolumn{16}{l}{\textcolor{t2e_green}{$\bullet$~\textbf{Type: Open-Source General Model}}}\\

Llama-3.2-90B-Vision-Ins~\cite{llama3.2} 
& 90B & 32.00 & 38 & 34 & 44 & 28 & 32 & 16 
& 40.30 & \ulshade{94} & 24 & 50 & 32 & 28 & 14 \\

InternVL2.5-8B~\cite{internvl2.5} 
& 8B & 2.00 & 4 & 6 & 2 & 0 & 0 & 0 
& 11.30 & 36 & 4 & 0 & 10 & 16 & 2 \\

Gemma-3-12b-its~\cite{gemma3technicalreport} 
&12B & 25.70 & 32 & 26 & 38 & 26 & 20 & 12 
& 23.00 & 58 & 10 & 24 & 18 & 24 & 4 \\

Qwen2.5-VL-72B-Ins$^{\ddagger}$~\cite{qwen2.5vl} 
& 72B & 43.67 & 62 & 36 & 48 & 40 & 44 & 32 
& 50.33 & 92 & 38 & 48 & 34 & 46 & 44 \\

Qwen2.5-VL-7B-Ins$^{\ddagger}$~\cite{qwen2.5vl} 
& 7B& 2.67 & 6 & 2 & 6 & 0 & 2 & 0 
& 15.00 & 42 & 6 & 22 & 12 & 4 & 4 \\

Qwen2.5-VL-3B-Ins$^{\ddagger}$~\cite{qwen2.5vl} 
& 3B & 1.33 & 2 & 2 & 0 & 0 & 4 & 0 
& 14.67 & 34 & 0 & 18 & 18 & 14 & 4 \\

\rowcolor{t2e_red!15}\multicolumn{16}{l}{\textcolor{t2e_red}{$\bullet$~\textbf{Type: Embodied Planning Model}}}\\

RoboBrain$^{\dagger}$~\cite{robobrain} 
& 7B & 0.33 & 2 & 0 & 0 & 0 & 0 & 0 
& 15.30 & 38 & 6 & 18 & 8 & 18 & 4 \\

Tapa$^{\dagger}$~\cite{TAPA} 
& 7B & 0.00 & 0 & 0 & 0 & 0 & 0 & 0 
& 0.00 & 0 & 0 & 0 & 0 & 0 & 0 \\

REBP $^{\ddagger}$~\cite{REBP}   
& 7B & 35.00 & 52 & 46 & 46 & 28 & 32 & 6  
& 18.33 & 50 & 6 & 18 & 14 & 14 & 8 \\

\rowcolor{yellow!18}
\textsf{RoboGPT-R1 (ours)}$^{\ddagger}$ 
& \textbf{3B}& 55.33 & 62 & 56 & 64 & 50 & 50 & 50 
& 22.00 & 64 & 8 & 18 & 20 & 12 & 10 \\

\bottomrule
\end{tabular}}
\end{table*}

\noindent\textbf{Dataset.}
We process the REBP public dataset~\cite{REBP} to generate a base dataset and an augmented dataset. The base dataset is directly distilled from the EB-ALFRED tasks in EmbodiedBench~\cite{EmbodiedBench}, serving as a benchmark-aligned dataset used in the SFT phase to endow the model with initial multimodal embodied-planning skills. The main body of the augmented dataset originates from the open-source ALFRED trajectory dataset~\cite{ALFRED}. While its content is similar to EB-ALFRED, it exhibits significant differences in details such as action space, visual appearance, task types, and task length, making it a benchmark-adjacent (near-domain) dataset. The augmented dataset includes all embodied planning data in the base dataset to prevent catastrophic forgetting. It is employed in the RFT phase to improve reasoning robustness under near-domain distributions, thereby further enhancing planning performance on EmbodiedBench. 
Data processing methods, data composition, and other details are provided in the appendix~\ref{app:data}.

\noindent\textbf{Training.}
We adopt \texttt{Qwen2.5-VL-3B-Instruct} as the multimodal base model and employ a two-stage training scheme. 
In the first stage, we perform full-parameter SFT on the \emph{base} dataset to endow the model with initial planning skills aligned with \emph{EB-ALFRED}. 
In the second stage, we conduct reinforcement fine-tuning (RFT) with GRPO on the \emph{augmented} dataset, aiming to improve reasoning and generalization from data that are not strictly benchmark-matched. 
SFT is implemented with \texttt{LLaMA-Factory}~\cite{llamafactory} and trained on 8$\times$ Ascend~910B3 64GB NPUs for about 1.5\,hours; 
RFT is implemented with \texttt{VERL}~\cite{easyr1} and trained on 4$\times$ NVIDIA H20 96GB GPUs for about 25\,hours. 
Complete hyperparameters and implementation details are provided in the appendix~\ref{app:exp}.

\begin{figure}[t]
  \centering
  \begin{minipage}[t]{0.49\linewidth}
    \centering
    \includegraphics[width=\linewidth]{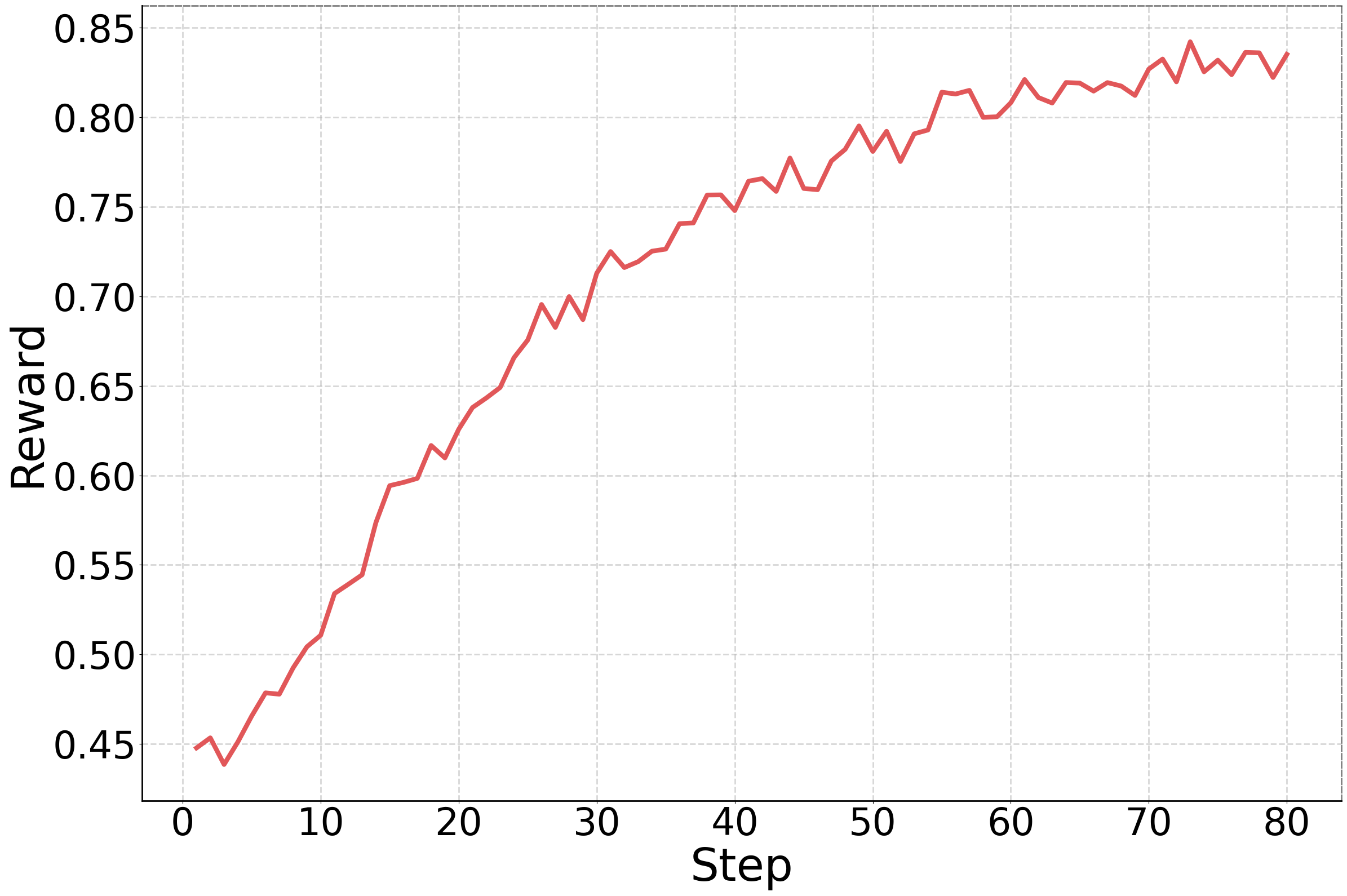}
    \vspace{2pt}
    {\footnotesize (a) Overall Reward}
  \end{minipage}\hfill
  \begin{minipage}[t]{0.49\linewidth}
    \centering
    \includegraphics[width=\linewidth]{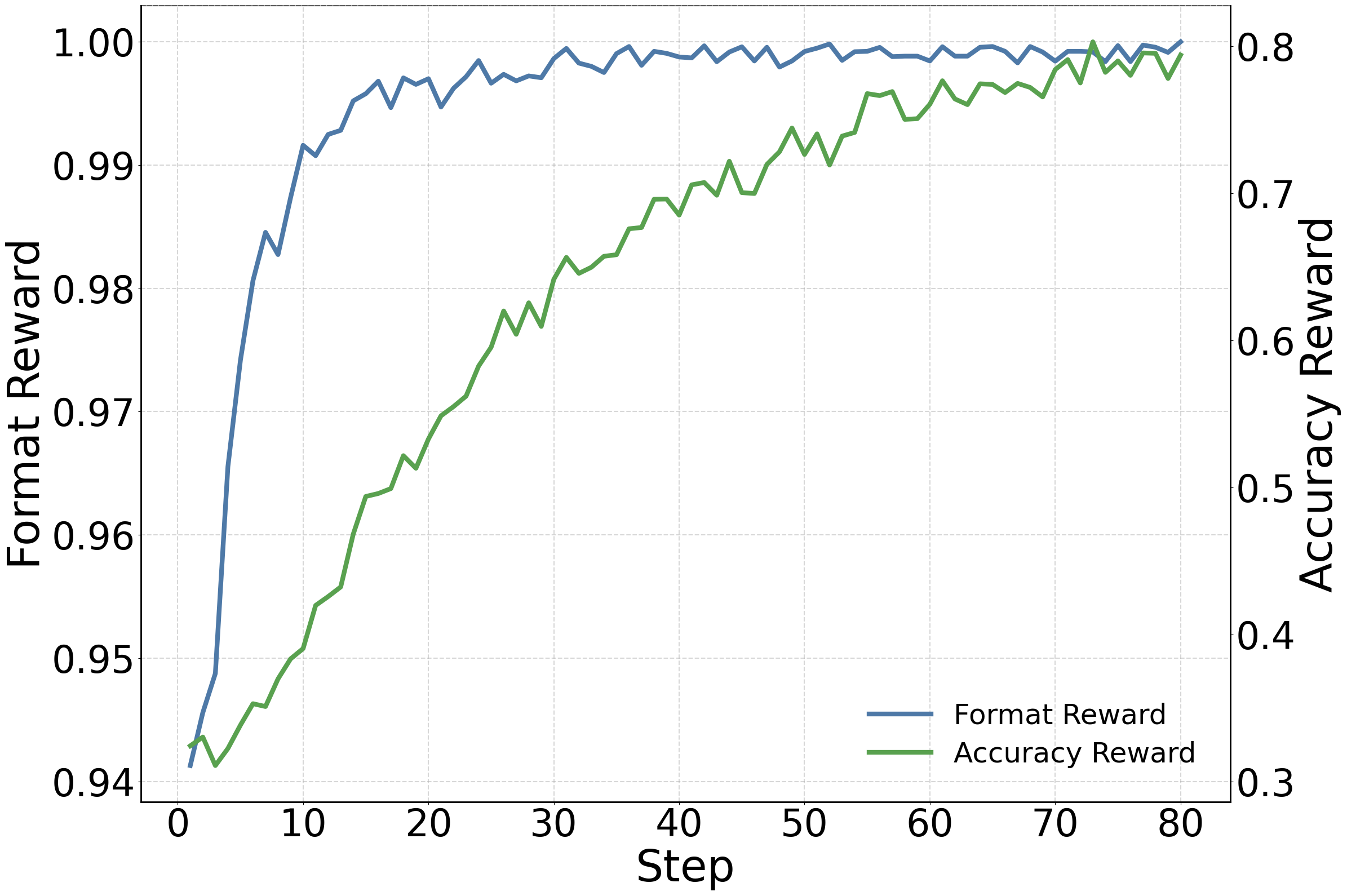}
    \vspace{2pt}
    {\footnotesize (b) Format \& Accuracy(LCS) Reward}
  \end{minipage}
  \Description{Two line plots showing reward curves over training steps during RFT. Panel (a) plots the overall reward increasing steadily with step. Panel (b) plots two curves: the format reward quickly rises and saturates near 1.0, while the accuracy (LCS) reward increases more gradually toward about 0.8.}
  \caption{Reward Curves in RFT. Our LCS-based accuracy reward provides an appropriate and dense learning signal for RFT, rising steadily from ~0.30 to ~0.80. The format reward, already aligned by SFT, starts around 0.95 at the onset of RFT and saturates within $\sim$20 steps, stabilizing near 1.0. The overall reward is computed as a weighted sum of the accuracy and format rewards with weights 0.8 and 0.2, respectively, and it increases in tandem with the steady improvement of the accuracy reward. }
  \label{fig:reward-curves}
\end{figure}


\begin{table*}[!t]
    \centering
    \caption{Training-Strategy Ablation Result (success rate, \%). After SFT, the model acquires core embodied planning competence, raising the average success rate from 1.33\% to 42.00\% while still lagging on long-horizon tasks. With subsequent RFT, performance further increases to 55.33\%, with especially pronounced gains on long-horizon tasks (26\% → 50\%).}
    \label{tab:ablation_training}
    \resizebox{0.85\linewidth}{!}{
    \begin{tabular}{ccc|l|llllll}
    \toprule
    \textbf{Base} & \textbf{SFT} & \textbf{RFT} & \textbf{Avg.}& \textbf{Base} & \textbf{Common} &\textbf{Complex} &
    \textbf{Visual} & \textbf{Spatial} & \textbf{Long}
    \\\midrule\midrule
    \textcolor{t2e_red}{\cmark}& \textcolor{t2e_gray}{\xmark} & \textcolor{t2e_gray}{\xmark} & $1.33$\textcolor{gray}{$_{(+0.00)}$} & 2\textcolor{gray}{$_{(+0.00)}$}  & 2\textcolor{gray}{$_{(+0.00)}$}  & 0\textcolor{gray}{$_{(+0.00)}$}  & 0\textcolor{gray}{$_{(+0.00)}$}  & 4\textcolor{gray}{$_{(+0.00)}$}  & 0 \textcolor{gray}{$_{(+0.00)}$}
    \\
    
    \textcolor{t2e_red}{\cmark} & \textcolor{t2e_blue}{\cmark} & \textcolor{t2e_gray}{\xmark} & $42.00$\textcolor{t2e_blue}{$_{(+41.33)}$} & 48\textcolor{t2e_blue}{$_{(+46)}$} & 44\textcolor{t2e_blue}{$_{(+42)}$} & 58 \textcolor{t2e_blue}{$_{(+58)}$}& 38 \textcolor{t2e_blue}{$_{(+38)}$}& 38\textcolor{t2e_blue}{$_{(+34)}$} & 26\textcolor{t2e_blue}{$_{(+26)}$}

    \\\midrule

    \textbf{\textcolor{t2e_red}{\cmark}} & \textbf{\textcolor{t2e_blue}{\cmark}} & \textbf{\textcolor{t2e_green}{\cmark}} & $\textbf{55.33}$\textbf{\textcolor{t2e_red}{$_{(+54.00)}$}} & \textbf{62\textcolor{t2e_red}{$_{(+60)}$}} & \textbf{56\textcolor{t2e_red}{$_{(+54)}$}} & \textbf{64\textcolor{t2e_red}{$_{(+64)}$}} & \textbf{50\textcolor{t2e_red}{$_{(+50)}$}} & \textbf{50\textcolor{t2e_red}{$_{(+46)}$}} & \textbf{50\textcolor{t2e_red}{$_{(+50)}$}}

    \\
    \bottomrule
    \end{tabular}}
 
\end{table*}


\begin{table*}[!t]
  \caption{Data-Source Ablation Result (success rate, \%). Base denotes the in-domain dataset distilled from EMbench; Aug is the near-domain ALFRED-derived set. Training only with SFT on Base reaches 40.00\% on average, whereas replacing Base with Aug for SFT collapses performance to 6.00\%, indicating poor transfer under pure supervision. Continuing RFT on Base after SFT on Base yields only a modest gain (40.00\% → 44.33\%), while using Aug during RFT achieves the best results (55.33\%), showing that only the RL (RFT) stage can effectively absorb near-domain data and transfer it to the target task.}
  \label{tab:ablation_dataset}
  \centering
  \resizebox{0.85\linewidth}{0.08\textheight}{%
  \begin{tabular}{l| l| c c c c c c}
    \toprule
    \textbf{Model} & \textbf{Avg.} & \textbf{Base} & \textbf{Common} &
    \multicolumn{1}{c}{\textbf{\shortstack{Complex}}} &
    \textbf{Visual} & \textbf{Spatial} & \textbf{Long} \\
    \midrule\midrule
    Base Model (Qwen2.5-VL-3B)         &  1.33 &  2 &  2 &  0 &  0 &  4 &  0 \\
    \midrule
    Only SFT w Base (ours)             & 42.00 & 48 & 44 & 58 & 38 & 38 & 26 \\
    Only SFT w Aug                    &  6.00 & 14 &  6 & 12 &  2 &  2 &  0 \\
    \midrule
    SFT+ RFT w Base                   & 44.33 & 56 & 56 & 54 & 32 & 36 & 32 \\
    SFT+ RFT w Aug (ours)              & 55.33 & 62 & 56 & 64 & 50 & 50 & 50 \\
    \bottomrule
  \end{tabular}}
\end{table*}

\subsection{Main Results}
\noindent\textbf{Training Results in EB-ALFRED.} Table~\ref{tab:main_result} reports the evaluation results in EB-ALFRED. RoboGPT-R1 attains an average success rate of \textbf{55.33\%} across six sub-task suites. This performance \textbf{significantly outperforms} multiple strong baselines: it surpasses the closed-source \textbf{GPT-4o (51.67\%)} and \textbf{GPT-4o-mini (34.00\%)}, and trails only \textbf{GPT-4.1 (64.67\%)}. Among open-source general models, it notably outperforms the larger-parameter \textbf{Qwen2.5-VL-72B-Instruct (43.67\%)}. Relative to the small-scale model with similar parameters \textbf{Qwen2.5-VL-3B-Instruct}, our approach yields an approximately \textbf{54\%} relative improvement in average success. Compared with the embodied specialist \textbf{REBP (35.00\%)}, RoboGPT-R1 leads by nearly \textbf{20} percentage points overall and shows a pronounced advantage on long-horizon tasks: REBP achieves only \textbf{6\%}, whereas RoboGPT-R1 reaches \textbf{50\%}. Notably, despite using only \textbf{3B} parameters, RoboGPT-R1 delivers this level of performance under a small-model, low-inference-cost setting, highlighting the parameter efficiency and effectiveness of our method.

\noindent\textbf{Generalization Results in EB-Habitat.} Table~\ref{tab:main_result} summarizes the \textbf{EB-Habitat} results. \textbf{RoboGPT-R1} attains an average success rate of 22\% across six sub-task suites, representing a 7\% improvement over the base model \textbf{Qwen2.5-VL-3B-Instruct} (14.67\%) and outperforming both the 7B-parameter models \textbf{Qwen2.5-VL-7B-Instruct} (15.00\%) and \textbf{REBP} (18.33\%). Although a performance gap remains compared to \textbf{Qwen2.5-VL-72B-Instruct} and several closed-source general models, these \emph{out-of-domain} results indicate that our approach substantially improves generalization and transferability of the embodied model.

\subsection{Ablation Study}

We conduct two sets of ablations on \textbf{EB-ALFRED} to disentangle the effects of the two training stages and the two data sources used in our framework. For brevity, we refer to the \emph{Base dataset} and the \emph{Aug dataset} as \textbf{Base} and \textbf{Aug}, respectively.

\begin{figure}[t]
  \centering
  \includegraphics[width=\linewidth]{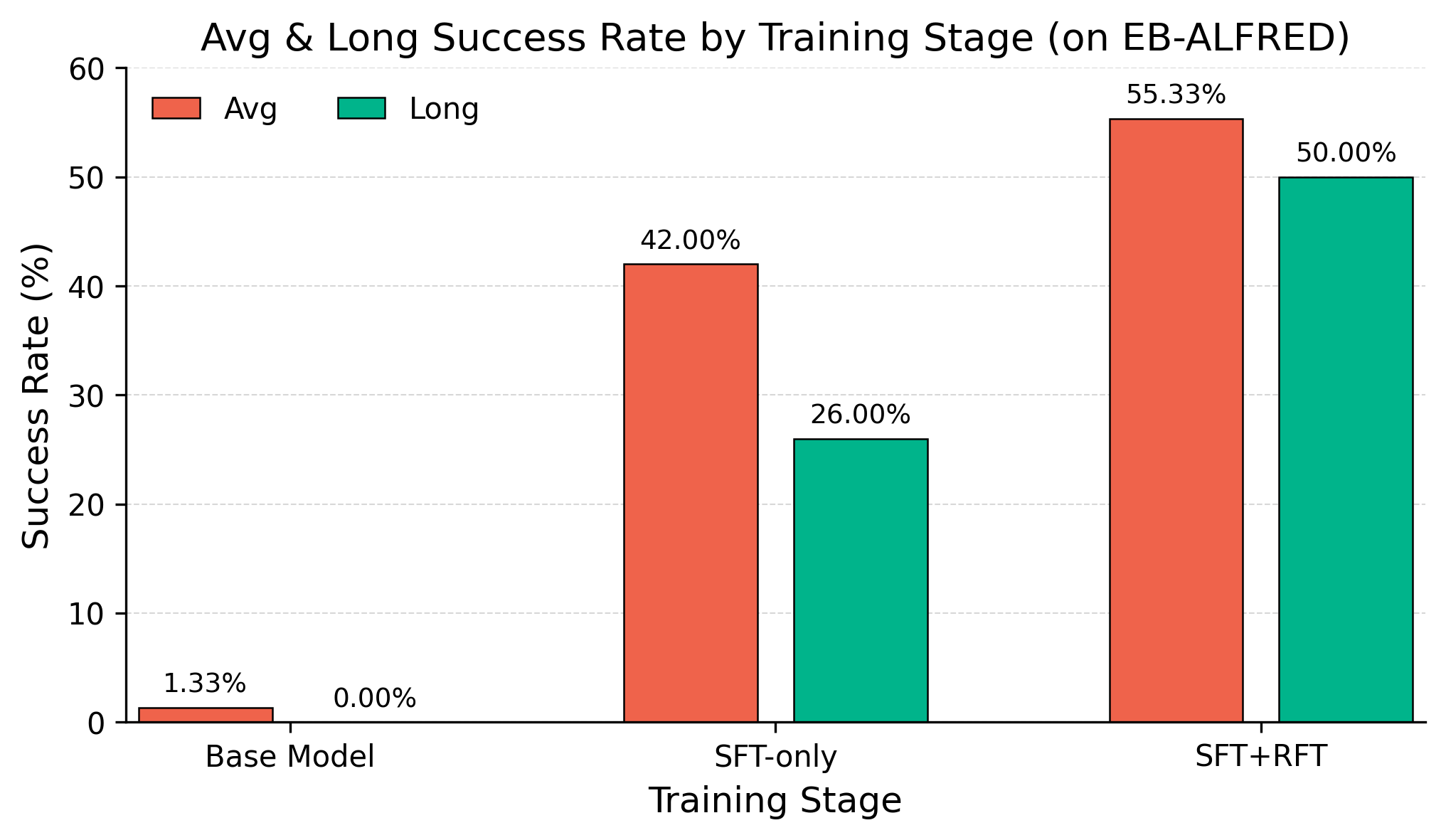} 
  \Description{Bar chart titled “Avg & Long Success Rate by Training Stage (on EB-ALFRED)”. It compares three models (Base Model, SFT-only, and SFT+RFT) with two bars per model (Avg and Long), showing success rates increasing from near zero for the base model to higher values after SFT and highest after SFT+RFT.}
  \caption{Success rates with different stages. Bars show the macro average and the long-horizon score for the base model, SFT-only, and SFT+RFT model. SFT establishes initial embodied planning (Avg: 1.33\%→42.00\%) but leaves long-horizon performance limited (26\%). Adding RFT lifts the averages further (to 55.33\%) and markedly improves long-horizon planning (to 50\%), validating the effectiveness of our two-stage framework—SFT for foundational competence and RFT for the additional gains needed to solve long-horizon tasks.}
  \label{fig:success_rate_by_stage}
\end{figure}

\noindent\textbf{Training-Strategy Ablation.}
We compare the performance changes on \emph{EB-ALFRED} among the base model (no fine-tuning), the SFT-only model, and the final SFT+RFT model, as shown in Table~\ref{tab:ablation_training} and Fig.~\ref{fig:success_rate_by_stage}. After SFT, the average success rate rises from \textbf{1.33\%} (base model) to \textbf{42.00\%}, indicating that SFT learns the initial embodied planning competence. However, performance on long-horizon tasks remains limited (\textbf{26\%}). With subsequent RFT, performance improves across all sub-task suites, yielding an average of \textbf{55.33\%}; notably, the long-horizon score increases from \textbf{26\%} to \textbf{50\%}, highlighting the particular effectiveness of RFT for complex, extended plans.

\begin{table*}[!t]
\centering
\caption{Accuracy Reward Comparison Result (success rates, \%). We compare three GRPO accuracy rewards used during RFT—Step Accuracy, REBP Acc., and LCS (ours)—with “RFT Base” as the reference row for deltas. Bracketed scores denote the change relative to the reference. LCS (ours) delivers the strongest overall performance (Avg. 55.33\%) and the largest increase on long-horizon tasks (+24\%), outperforming the alternatives across most sub-suites.}
\label{tab:acc_reward_ablation}
\setlength{\tabcolsep}{2pt}
\renewcommand{\arraystretch}{1.05}
\resizebox{0.8\linewidth}{0.065\textheight}{%
\begin{tabular}{l|l|llllll}
\toprule
\textbf{Accuracy Reward} & \textbf{Avg.} & \textbf{Base} & \textbf{Common} & \textbf{Complex} & \textbf{Visual} & \textbf{Spatial} & \textbf{Long} \\
\midrule\midrule
RFT Base
  & 42.00\textcolor{gray}{\scriptsize\, (+0.00)}
  & 48\textcolor{gray}{\scriptsize\, (+0.00)}
  & 44\textcolor{gray}{\scriptsize\, (+0.00)}
  & 58\textcolor{gray}{\scriptsize\, (+0.00)}
  & 38\textcolor{gray}{\scriptsize\, (+0.00)}
  & 38\textcolor{gray}{\scriptsize\, (+0.00)}
  & 26\textcolor{gray}{\scriptsize\, (+0.00)} \\
Step Accuracy
  & 43.67\textcolor{t2e_blue}{\scriptsize\, (+1.67)}
  & 52\textcolor{t2e_blue}{\scriptsize\, (+4.00)}
  & 56\textcolor{t2e_blue}{\scriptsize\, (+12.00)}
  & 54\textcolor{t2e_blue}{\scriptsize\, (-4.00)}
  & 40\textcolor{t2e_blue}{\scriptsize\, (+2.00)}
  & 40\textcolor{t2e_blue}{\scriptsize\, (+2.00)}
  & 20\textcolor{t2e_blue}{\scriptsize\, (-6.00)} \\
REBP Acc.
  & 48.33\textcolor{t2e_blue}{\scriptsize\, (+6.33)}
  & 62\textcolor{t2e_blue}{\scriptsize\, (+14.00)}
  & 52\textcolor{t2e_blue}{\scriptsize\, (+8.00)}
  & 58\textcolor{t2e_blue}{\scriptsize\, (+0.00)}
  & 46\textcolor{t2e_blue}{\scriptsize\, (+8.00)}
  & 38\textcolor{t2e_blue}{\scriptsize\, (+0.00)}
  & 34\textcolor{t2e_blue}{\scriptsize\, (+8.00)} \\
\textbf{LCS (ours)}
  & \textbf{55.33}\textcolor{t2e_red}{\scriptsize\, (+13.33)}
  & \textbf{62}\textcolor{t2e_red}{\scriptsize\, (+14.00)}
  & \textbf{56}\textcolor{t2e_red}{\scriptsize\, (+12.00)}
  & \textbf{64}\textcolor{t2e_red}{\scriptsize\, (+8.00)}
  & \textbf{50}\textcolor{t2e_red}{\scriptsize\, (+12.00)}
  & \textbf{50}\textcolor{t2e_red}{\scriptsize\, (+12.00)}
  & \textbf{50}\textcolor{t2e_red}{\scriptsize\, (+24.00)} \\
\bottomrule
\end{tabular}}
\end{table*}

\noindent\textbf{Data-Source Ablation.}
As shown in Table~\ref{tab:ablation_dataset}, these ablations highlight the importance of near-domain data during RFT.
(1) \emph{SFT on Aug only:} Replacing Base with Aug for SFT reduces the average success from \textbf{42.00\%} to \textbf{6.00\%}, suggesting that, under supervised learning, near-domain data alone does not transfer effectively to the target task.
(2) \emph{RFT on Base:} Starting from SFT on Base and continuing RFT on Base (instead of Aug) yields only a modest increase from \textbf{42.00\%} to \textbf{44.33\%}. Compared with the SFT+RFT setting that uses Aug during RFT (\textbf{55.33\%}), this indicates that incorporating near-domain data in the RFT stage is essential.

\noindent\textbf{Ablation Analysis.}
We summarize three observations: (i) SFT establishes the initial planning ability but leaves a gap on long-horizon tasks; (ii) RFT closes this gap with the strongest gains on long-horizon sub-task suites; (iii) combining RFT with Aug is essential, as Aug brings little benefit in SFT but provides substantial gains when used during RFT.

\subsection{Accuracy Reward Comparison}
\label{sec:acc-reward}

During the RFT stage, we instantiate the accuracy reward in the GRPO algorithm~\cite{grpo} as the normalized \emph{Longest Common Subsequence (LCS)} length between the generated and reference trajectories. Thanks to the appropriate and dense learning signal provided by the LCS-based accuracy reward, the reward curves rise steadily throughout RFT, as shown in Fig.~\ref{fig:reward-curves}. This section conducts a head-to-head comparison of three \emph{accuracy rewards} to assess the suitability and advantages of our LCS reward for embodied planning.

\noindent\textbf{Compared rewards.}
We set the accuracy reward in GRPO to one of:
\begin{itemize}
  \item \textbf{LCS reward (ours):} the ratio of the longest common subsequence between the generated and reference trajectories (Normalized LCS);
  \item \textbf{Step Accuracy:}  the ratio of strictly matched actions/instructions step-by-step;  
  \item \textbf{REBP reward:} the multi-step, progress-style signal used in REBP~\cite{REBP}.
\end{itemize}

We evaluate on \textbf{EB-ALFRED} with the same settings as in the main experiments. To ensure fairness, we fix the backbone model, training data, total update budget, optimization hyperparameters, and the format reward term, and vary only the accuracy reward. The primary metrics are the \emph{average success rate} and the scores of each sub-task suite (with emphasis on \emph{long-horizon} tasks). 

\noindent\textbf{Comparison Results.}
As summarized in Table~\ref{tab:acc_reward_ablation}, using \textbf{Step Accuracy} yields only a slight improvement in the average success rate (from \textbf{42.00\%} to \textbf{43.67\%}), while the long-horizon score decreases (from \textbf{26\%} to \textbf{20\%}). The \textbf{REBP reward} improves the average to \textbf{48.33\%} and raises the long-horizon score from \textbf{26\%} to \textbf{34\%}. In contrast, our \textbf{LCS reward} achieves the largest gains: the average increases from \textbf{42.00\%} to \textbf{55.33\%}, and the long-horizon score from \textbf{26\%} to \textbf{50\%}. These findings demonstrate the suitability of the LCS-based accuracy reward for embodied planning and its advantages over the alternatives under the same training budget.

\section{CONCLUSIONS}
\label{sec:conclusions}

In this work, we propose RoboGPT-R1, a two-stage training framework for embodied planning. Stage 1 (SFT) equips the model with initial instruction-following and planning priors. Stage 2 (RFT) performs reinforcement fine-tuning with GRPO and an LCS-based, accuracy reward (paired with format constraints), providing dense and verifiable feedback. This design overcomes the limitations of SFT-only behavior cloning, which fails to adequately elicit the reasoning capabilities of VLMs and often undermines in-domain performance when leveraging near-domain data. These shortcomings result in poor performance on long-horizon tasks and brittle out-of-domain generalization. Evaluated on EmbodiedBench, the 3B-parameter model trained with our framework surpasses general-purpose VLM baselines such as Qwen2.5-VL-72B and GPT-4o, and substantially outperforms other 7B-scale embodied planners, with especially pronounced gains on long-horizon subtasks.


\begin{acks}

This work was supported by the National Natural Science Foundation of China (NSFC) under Grants 62136008 and 62293545; in part by
the Beijing Major Science and Technology Project under Contract No.
Z251100008125023; and by the Suzhou Innovation and Entrepreneurship
Leading Talents Programe – Innovation Leading Talent in Universities and
Research Institutes under Grant ZXL2025310; and by Beijing Academy of
Artificial Intelligence (BAAI).
\end{acks}


\balance






\bibliographystyle{ACM-Reference-Format} 
\bibliography{sample}


\clearpage
\onecolumn
\appendix

\begin{center}
    \LARGE \textbf{Appendix}
\end{center}

\section{EXPERIMENTAL DETAILS}\label{app:exp}

\subsection{SFT Details}

In all experiments conducted in this paper, the hyperparameter settings for the supervised fine-tuning (SFT) phase strictly adhere to the configurations listed in the table~\ref{tab:hpp_sft}, using 8$\times$ Ascend~910B3 64GB NPUs as computational devices.  We empirically set the parameter  \texttt{"num\_train\_epochs"} to 2 epochs in all SFT experiments to ensure that the model learns effectively without severe overfitting.


\begin{table*}[htbp]
\centering
\caption{SFT hyperparameter settings used in our experiments.}
\setlength{\tabcolsep}{2pt}
\renewcommand{\arraystretch}{1.0}

\begin{tabular}{@{} l r | l r @{}}
\toprule
\textbf{Hyperparameter} & \textbf{Value} &
\textbf{Hyperparameter} & \textbf{Value} \\
\midrule

image\_max\_pixels & 262144 & cutoff\_len & 9216 \\ 
video\_max\_pixels & 16384 & max\_samples & 50000 \\ 
trust\_remote\_code &\texttt{true} & overwrite\_cache &\texttt{true} \\ 
stage &\texttt{sft} & per\_device\_train\_batch\_size & 1 \\ 
finetuning\_type &\texttt{full} & gradient\_accumulation\_steps & 2 \\ 
freeze\_vision\_tower &\texttt{true} & learning\_rate & 0.00001 \\ 
freeze\_multi\_modal\_projector &\texttt{true} & num\_train\_epochs & 2.0 \\ 
freeze\_language\_model &\texttt{false} & lr\_scheduler\_type &\texttt{cosine} \\ 
deepspeed & ds\_z2\_config.json & warmup\_ratio & 0.1 \\ 
ddp\_timeout & 180000000 & bf16 &\texttt{true} \\ 
template &\texttt{qwen2\_vl} & nproc\_per\_node & 8 \\  

\bottomrule
\end{tabular}

\label{tab:hpp_sft}
\end{table*}

\subsection{RFT Details}
The hyperparameter settings remained consistent across all experiments during the RFT phase, as listed in Table~\ref{tab:hpp_rft}. The RFT phase utilized 4$\times$ NVIDIA H20 96GB GPUs as computational devices. For the sake of fairness in comparison, the iteration count for the RFT phase was uniformly set to 80 steps.

\begin{table*}[t]
\centering
\caption{RFT hyperparameter settings used in our experiments.}
\small
\setlength{\tabcolsep}{2pt}
\renewcommand{\arraystretch}{1.0}

\begin{tabular}{@{} l r | l r @{}}
\toprule
\textbf{Hyperparameter} & \textbf{Value} &
\textbf{Hyperparameter} & \textbf{Value} \\
\midrule

\multicolumn{2}{>{\columncolor{t2e_blue!15}}c|}{\textcolor{t2e_blue}{\textbf{data}}} &
actor\_rollout\_ref.ref.log\_prob\_use\_dynamic\_bsz & \texttt{true} \\
data.video\_fps & 2.0 & actor\_rollout\_ref.ref.log\_prob\_max\_token\_len\_per\_gpu & 16384 \\
data.min\_pixels & 40000 & actor\_rollout\_ref.ref.ulysses\_sequence\_parallel\_size & 1 \\
data.max\_pixels & 4194304 & \multicolumn{2}{>{\columncolor{t2e_blue!15}}c}{\textcolor{t2e_blue}{\textbf{actor\_rollout\_ref.rollout}}} \\
data.max\_prompt\_length & 6144 & actor\_rollout\_ref.rollout.name & \texttt{vllm} \\
data.max\_response\_length & 3072 & actor\_rollout\_ref.rollout.temperature & 1.0 \\
data.train\_batch\_size & 512 & actor\_rollout\_ref.rollout.top\_k & -1 \\
data.shuffle & \texttt{true} & actor\_rollout\_ref.rollout.top\_p & 1.0 \\
\multicolumn{2}{>{\columncolor{t2e_blue!15}}c|}{\textcolor{t2e_blue}{\textbf{actor\_rollout\_ref.actor}}} &
actor\_rollout\_ref.rollout.n & 5 \\
actor\_rollout\_ref.actor.ppo\_mini\_batch\_size & 128 & actor\_rollout\_ref.rollout.response\_length & 3072 \\
actor\_rollout\_ref.actor.ppo\_micro\_batch\_size\_per\_gpu & 1 & actor\_rollout\_ref.rollout.gpu\_memory\_utilization & 0.6 \\
actor\_rollout\_ref.actor.use\_dynamic\_bsz & \texttt{true} & actor\_rollout\_ref.rollout.ignore\_eos & \texttt{false} \\
actor\_rollout\_ref.actor.ppo\_max\_token\_len\_per\_gpu & 16384 & actor\_rollout\_ref.rollout.enforce\_eager & \texttt{false} \\
actor\_rollout\_ref.actor.grad\_clip & 1.0 & actor\_rollout\_ref.rollout.tensor\_model\_parallel\_size & 4 \\
actor\_rollout\_ref.actor.clip\_ratio & 0.2 & actor\_rollout\_ref.rollout.max\_num\_batched\_tokens & 12288 \\
actor\_rollout\_ref.actor.entropy\_coeff & 0.0 & actor\_rollout\_ref.rollout.max\_num\_seqs & 1024 \\
actor\_rollout\_ref.actor.use\_kl\_loss & \texttt{true} & \multicolumn{2}{>{\columncolor{t2e_blue!15}}c}{\textcolor{t2e_blue}{\textbf{algorithm}}} \\
actor\_rollout\_ref.actor.kl\_loss\_type & \texttt{low\_var\_kl} & algorithm.gamma & 1.0 \\
actor\_rollout\_ref.actor.kl\_loss\_coef & \texttt{1.0e-2} & algorithm.lam & 1.0 \\
actor\_rollout\_ref.actor.ppo\_epochs & 1 & algorithm.adv\_estimator & \texttt{grpo} \\
actor\_rollout\_ref.actor.shuffle & \texttt{false} & algorithm.use\_kl\_in\_reward & \texttt{false} \\
actor\_rollout\_ref.actor.ulysses\_sequence\_parallel\_size & 1 & algorithm.kl\_penalty & \texttt{low\_var\_kl} \\
actor\_rollout\_ref.actor.optim.lr & \texttt{1.0e-6} & algorithm.kl\_ctrl.type & \texttt{fixed} \\
actor\_rollout\_ref.actor.optim.lr\_warmup\_steps & -1 & algorithm.kl\_ctrl.kl\_coef & 0.001 \\
actor\_rollout\_ref.actor.optim.lr\_warmup\_steps\_ratio & 0.0 & algorithm.kl\_ctrl.horizon & 10000 \\
actor\_rollout\_ref.actor.optim.lr\_scheduler\_type & \texttt{constant} & algorithm.kl\_ctrl.target\_kl & 0.1 \\
actor\_rollout\_ref.actor.optim.total\_training\_steps & -1 & algorithm.rollout\_is & \texttt{false} \\
actor\_rollout\_ref.actor.optim.weight\_decay & \texttt{1.0e-2} & \multicolumn{2}{>{\columncolor{t2e_blue!15}}c}{\textcolor{t2e_blue}{\textbf{trainer}}} \\
\multicolumn{2}{>{\columncolor{t2e_blue!15}}c|}{\textcolor{t2e_blue}{\textbf{actor\_rollout\_ref.ref}}} &
trainer.nnodes & 1 \\
actor\_rollout\_ref.ref.log\_prob\_micro\_batch\_size\_per\_gpu & 2 & trainer.n\_gpus\_per\_node & 4 \\

\bottomrule
\end{tabular}

\label{tab:hpp_rft}
\end{table*}


\section{Dataset Details}\label{app:data}

\subsection{Embodied Planning Data Processing}
The embodied planning task data used in our constructed dataset was primarily processed based on two datasets released by REBP ~\cite{REBP}. We performed the following data processing steps: First, we cleaned and removed data from the original REBP dataset that contained obvious errors, such as incomplete responses or content that was abnormally truncated. Next, we addressed the inconsistency in the order of the four key-value pairs within the response content. Following the logical sequence of  "Perception → Reasoning → Providing Answers → Formatting Output", we standardized the sequence of key-value pairs  to: \texttt{"visual\_state\_description"}, \texttt{"reasoning\_and\_reflection"}, \texttt{"language\_plan"} and \texttt{"executable\_plan"}.

Additionally, the original data input contains a large amount of example information, which consums numerous tokens. We remove all example prompts to align the data to a zero-shot state. This configuration significantly reduces the data length, shortening the length of each data by approximately one-third. This, in turn, shortens model training time and reduces computing resource consumption during training.


\subsection{Composition of the Base and Aug Dataset}

\noindent\textbf{Base.} The Base dataset primarily consists of the REBP open-source SFT embodied planning data processed as described in the previous section, supplemented by a small amount of data extracted from the RoboVQA~\cite{Robovqa} and MATH-Vision~\cite{math} datasets. The final Base dataset contains over 5,000 samples, comprising over 4,000 embodied planning samples and over 1,000 for other tasks. The inclusion of other tasks aims to prevent the model from overfitting to embodied planning tasks, which could compromise its inherent multimodal perception and reasoning capabilities.


\noindent\textbf{Aug.} The core of the Aug dataset consists of over 40,000 entries processed from the REBP open-source RFT dataset following the methodology described in the previous section. Additionally, it incorporates all 4,000+ embodied planning data samples from the Base dataset to prevent catastrophic forgetting during the RFT phase. The final Aug dataset contains over 45,000 samples.

\section{EVALUATION DETAILS}\label{app:eval}
When evaluating models in EmbodiedBench~\cite{EmbodiedBench}, the number of input examples during evaluation can be controlled via the \texttt{"n\_shots"} parameter, with a default maximum setting of 10. As described in the previous section, we align all training data to the 0-shot state during the processing phase. Therefore, to maintain consistency with the training process, the final evaluation results were obtained using the setting \texttt{"n\_shots=0"}  when testing our model.

When testing general models, our preliminary experiments revealed that the n-shot strategy is indispensable for general models to complete EmbodiedBench tests. Furthermore, the number of examples provided significantly affects the performance of general models, as shown in Table~\ref{tab:n-shot-GPT}. When no examples are provided at all(that is, in 0-shot scenarios), both Qwen and GPT series models are completely unable to perform embodied planning within EB-ALFRED. Providing just 1 example versus 10 examples results in a significant gap in test performance. To ensure a conservative comparison, all results from the general models directly tested in this paper within the main results shown in Table~\ref{tab:main_result} were obtained under the setting \texttt{"n\_shots=10"}, i.e., using test results from the optimal performance configuration. This setting guarantees that the experimental results provide a conservative estimate of our method's performance advantage.

All other test parameters use the default settings of EmbodiedBench.

\begin{table}[htbp]
  \centering
  \caption{Impact of n-shots Strategy on the Performance of General Models on EB-ALFRED}
  \label{tab:n-shot-GPT}
  \begin{tabular}{lcccc}
    \toprule
    \textbf{model} & \textbf{0-shot} & \textbf{1-shot} & \textbf{10-shots} \\
    \midrule
    GPT-4.1     & 0     & 46.00 & 64.67 \\
    GPT-4o      & 0     & 34.33 & 51.67 \\
    GPT-4o-mini & 0     & 0     & 24.00 \\
    Qwen2.5-VL-72B-Ins.    & 0     & 35.00 & 43.67 \\
    Qwen2.5-VL-7B-Ins.     & 0     & 0.33  & 2.67  \\
    Qwen2.5-VL-3B-Ins.     & 0     & 0.67  & 1.33  \\
    \bottomrule
  \end{tabular}
\end{table}


\section{Failure Cases Study}
\label{app:case}
To better understand the RoboGPT-R1’s limitations and to provide guidance for future improvements, we conduct a brief qualitative inspection of a subset of failure cases. From these observations, we summarize several recurring failure behaviors and present representative examples with visualizations below.

\begin{figure}[h] 
  \centering
  \includegraphics[width=0.9\linewidth]{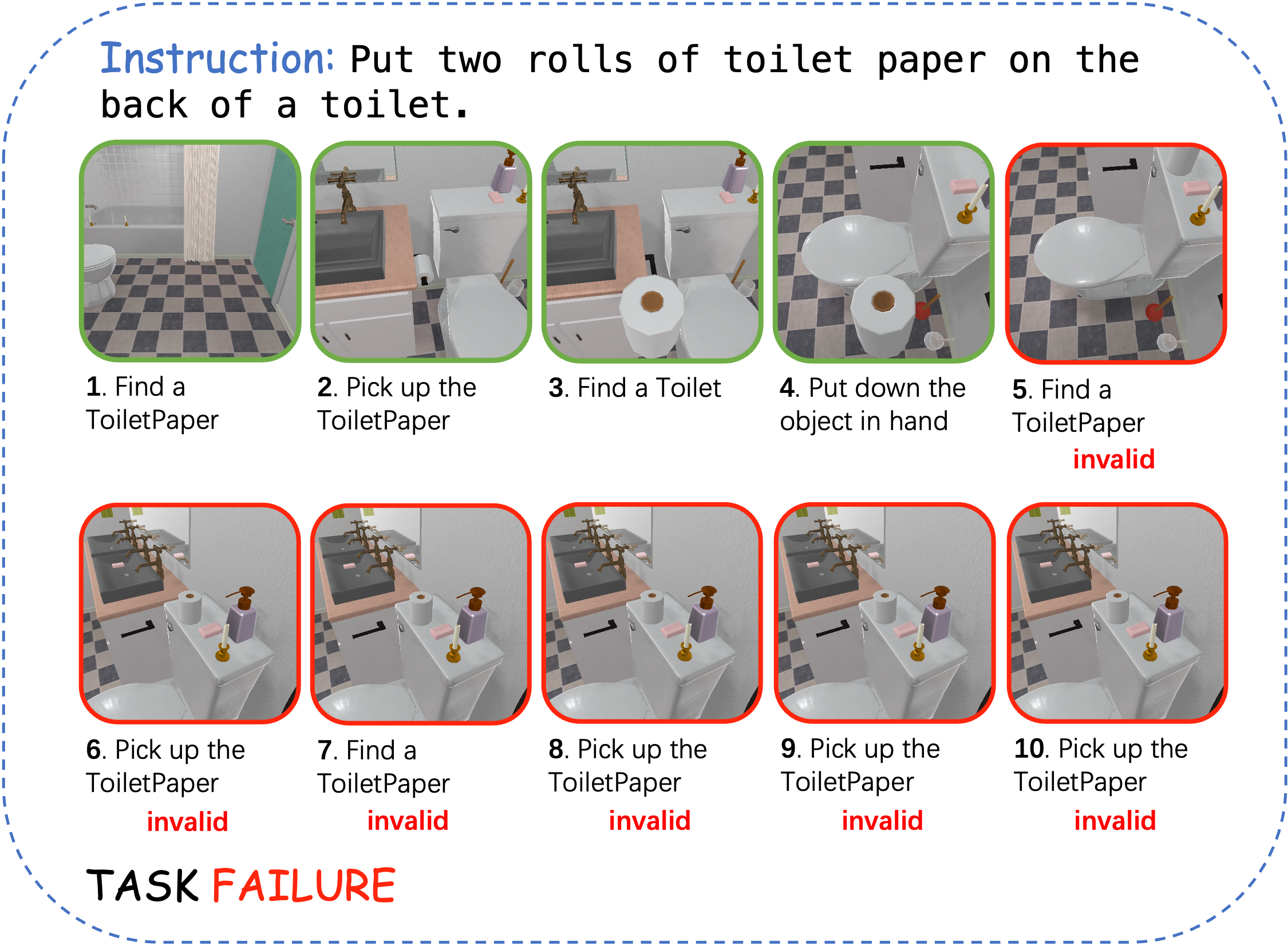}
  \Description{}
  \caption{Failure Case 1.}
  \label{fig:app_case1}
\end{figure}

\noindent\textbf{Case 1:} As shown in Fig.~\ref{fig:app_case1}, the agent successfully places the first roll of toilet paper onto the back of the toilet, but fails to complete the second placement. After the initial success, the already-placed roll becomes a salient visual distractor, and the agent repeatedly generates and executes a local loop of action plans such as \texttt{find} / \texttt{pick up} \texttt{ToiletPaper}. We observe that this type of error occurs more frequently in instructions involving two or more identical target objects. Overall, the case suggests that when an instance of the target class has been completed and remains visually prominent, the model may anchor its attention and goal to the finished instance and struggles to re-align the objective to the \emph{other} unfinished one, highlighting limitations in multi-instance disambiguation and cross-step state tracking.


\end{document}